%% file: main.tex
\PassOptionsToPackage{dvipsnames}{xcolor}

\documentclass[acmtog]{acmart}

\AtBeginDocument{%
  }

\usepackage{colortbl}
\usepackage{bm}

\usepackage{amssymb}
\usepackage{multirow}
\usepackage{placeins}

\copyrightyear{2026}
\acmYear{2026}
\setcopyright{cc}
\setcctype{by}
\acmConference[SIGGRAPH Conference Papers '26]{Special Interest Group on Computer Graphics and Interactive Techniques Conference Conference Papers}{July 19--23, 2026}{Los Angeles, CA, USA}
\acmBooktitle{Special Interest Group on Computer Graphics and Interactive Techniques Conference Conference Papers (SIGGRAPH Conference Papers '26), July 19--23, 2026, Los Angeles, CA, USA}
\acmDOI{10.1145/3799902.3811122}
\acmISBN{979-8-4007-2554-8/2026/07}

\begin{document}

\title[FreeOrbit4D]{FreeOrbit4D: Training-Free Arbitrary Camera Redirection for Monocular Videos via Foreground-Complete 4D Reconstruction}

\author{Wei Cao}
\email{weicao3@illinois.edu}
\orcid{0009-0005-5163-6484}
\affiliation{%
  \institution{University of Illinois Urbana-Champaign}
  \city{Champaign}
  \country{USA}
}

\author{Hao Zhang}
\email{haoz19@illinois.edu}
\orcid{0009-0009-2082-641X}
\affiliation{%
  \institution{University of Illinois Urbana-Champaign}
  \city{Urbana}
  \country{USA}
}

\author{Fengrui Tian}
\email{tianfr@upenn.edu}
\orcid{0000-0002-9577-5276}
\affiliation{%
  \institution{University of Pennsylvania}
  \city{Philadelphia}
  \country{USA}
}

\author{Yulun Wu}
\email{yulun5@illinois.edu}
\orcid{0009-0009-4982-9630}
\affiliation{%
  \institution{University of Illinois Urbana-Champaign}
  \city{Champaign}
  \country{USA}
}

\author{Yingying Li}
\email{yl101@illinois.edu}
\orcid{0000-0002-1858-4257}
\affiliation{%
  \institution{University of Illinois Urbana-Champaign}
  \city{Urbana}
  \country{USA}
}

\author{Shenlong Wang}
\email{shenlong@illinois.edu}
\orcid{0000-0002-7984-266X}
\affiliation{%
  \institution{University of Illinois Urbana-Champaign}
  \city{Urbana}
  \country{USA}
}

\author{Ning Yu}
\email{ningyu.hust@gmail.com}
\orcid{0009-0004-6865-1325}
\affiliation{%
  \institution{Eyeline Labs}
  \city{Los Angeles}
  \country{USA}
}

\author{Yaoyao Liu}
\authornote{Corresponding author.}
\email{lyy@illinois.edu}
\orcid{0000-0002-5316-3028}
\affiliation{%
  \institution{University of Illinois Urbana-Champaign}
  \city{Champaign}
  \country{USA}
}

\renewcommand{\shortauthors}{Cao et al.}

\input{sec/0_abstract}


\begin{teaserfigure}
  \centering
  \includegraphics[width=\linewidth]{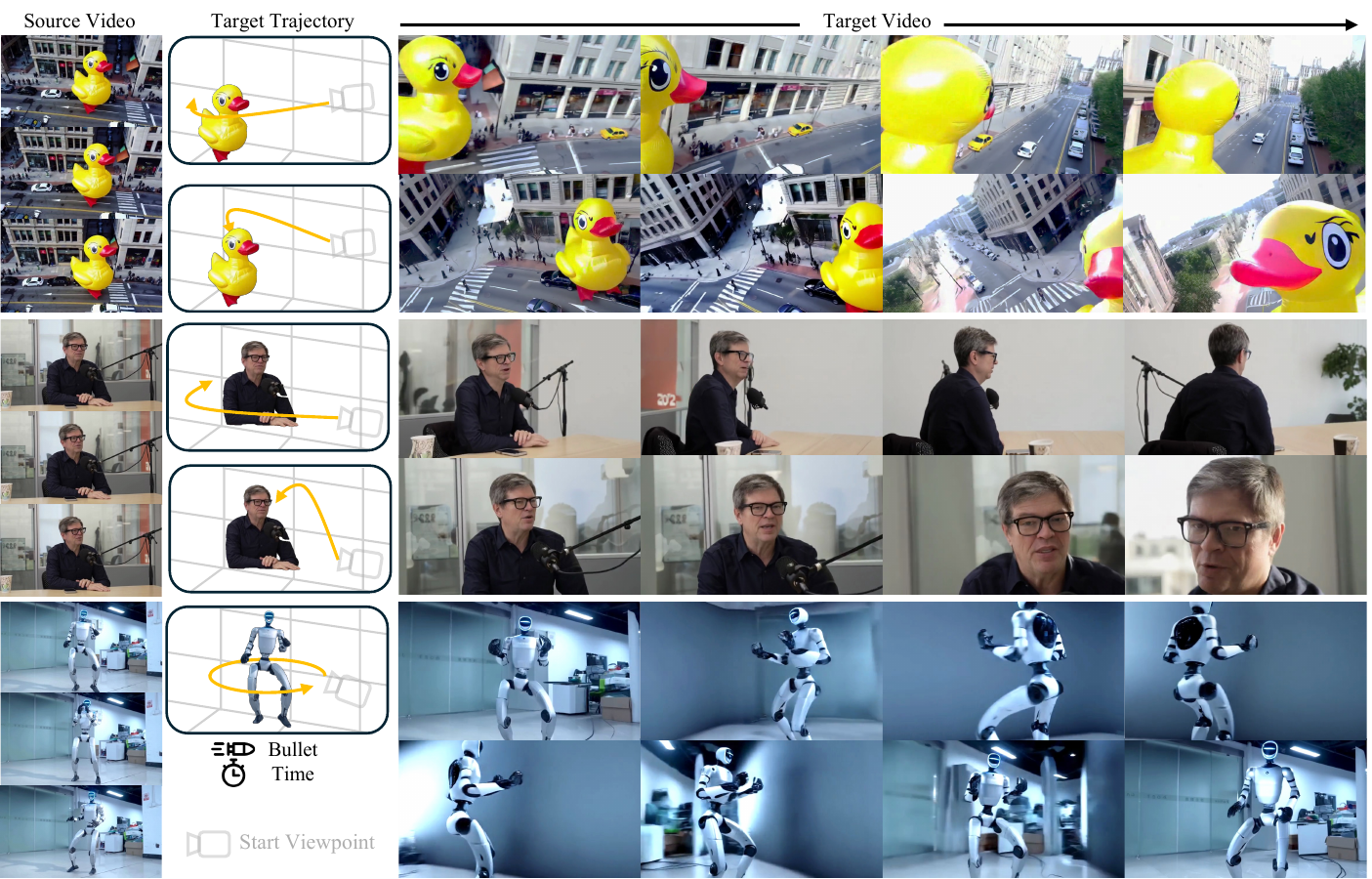}
  \caption{FreeOrbit4D enables \textbf{training-free} camera redirection from a single monocular video to arbitrary target camera trajectories. Given a source video (left) and a target trajectory (middle), our method produces a redirected video (right) with faithful appearance and strong temporal coherence under large-angle redirected camera motions, even including bullet-time orbits, demonstrated on diverse scenes and subjects.}
  \Description{A three-part horizontal teaser image showing the FreeOrbit4D pipeline. The left side shows a source monocular video frame; the middle shows a complex 3D target camera trajectory; the right side shows the resulting redirected video frames with consistent appearance and temporal coherence during a large-angle camera orbit.}
  \label{fig:teaser}
\end{teaserfigure}

\maketitle

\input{sec/1_introduction}
\input{sec/2_related_work}
\input{sec/3_method}
\input{sec/4_experiments}
\input{sec/5_ablation}
\input{sec/6_conclusion}

\begin{acks}
This research is supported by the National Artificial Intelligence Research Resource (NAIRR) Pilot under award NAIRR250199. Computational resources are also provided by Delta and DeltaAI at the National Center for Supercomputing Applications (NCSA) through ACCESS allocations CIS250012, CIS250816, and CIS251188. S. W. is also supported by NSF Awards \#2525287, \#2404385, \#2414227, \#2340254, \#2312102, and \#2331878, and research grants from IBM, Meta, NVIDIA, and Intel.
\end{acks}

\bibliographystyle{ACM-Reference-Format}
\bibliography{references}

\end{document}

%% file: sec/0_abstract.tex
\begin{abstract}

Camera redirection aims to replay a dynamic scene from a single monocular video under a user-specified camera trajectory.
However, large-angle redirection is inherently ill-posed: a monocular video captures only a narrow spatio-temporal view of a dynamic 3D scene, providing severely limited observations of the underlying 4D world.
The key challenge is therefore to recover a complete and coherent plenoptic representation from this limited input, with consistent geometry and coherent motion.
While recent diffusion-based methods achieve impressive visual generation quality, they often break down under large-angle viewpoint changes far from the original trajectory, where missing visual grounding leads to severe geometric ambiguity and temporal inconsistency.
To address this, we present \textbf{FreeOrbit4D}, an effective training-free framework that tackles this geometric ambiguity by recovering a foreground-complete 4D proxy as structural grounding for video generation.
We obtain this proxy by decoupling foreground and background reconstructions: we unproject the monocular video into a static background and partial foreground point clouds in a unified global space, then leverage an object-centric multi-view diffusion model to synthesize multi-view images and reconstruct complete foreground point clouds in canonical object space.
By aligning the canonical foreground point cloud to the global scene space via dense pixel-synchronized 3D--3D correspondences and projecting the foreground-complete 4D proxy onto target camera viewpoints, we provide geometric scaffolds (e.g., depth/visibility cues) that guide a conditional video diffusion model.
Extensive experiments show that FreeOrbit4D produces more faithful and temporally coherent redirected videos under challenging large-angle trajectories, and our foreground-complete 4D proxy further opens a potential avenue for practical applications such as edit propagation and 4D data generation.
\textbf{Project page:} \url{https://freeorbit4d.vision.ischool.illinois.edu/}

\end{abstract}

%% file: sec/1_introduction.tex
\section{Introduction}

\begin{figure}[!t]
  \centering
  \begin{minipage}{\linewidth}
    \centering
    \includegraphics[width=\linewidth]{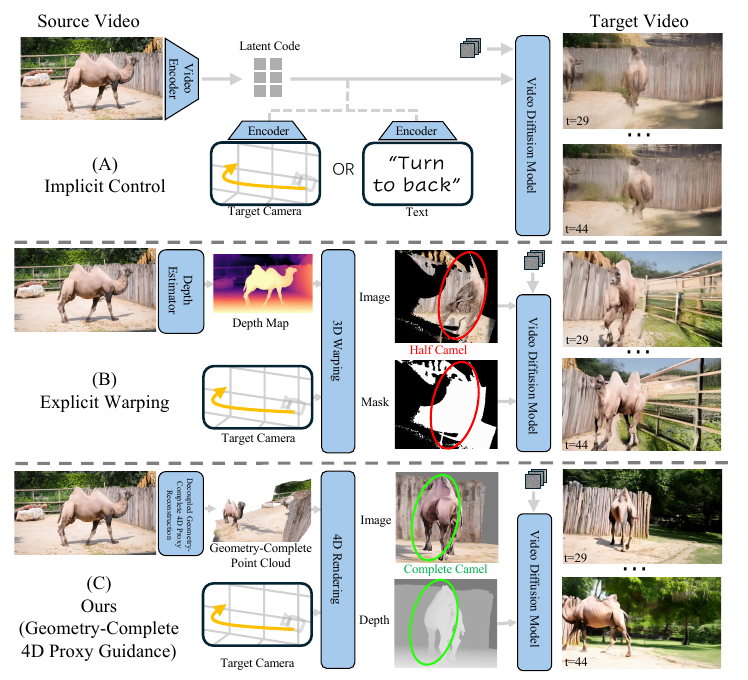}
  \end{minipage}
  \begin{minipage}{\linewidth}
    \centering
    \input{tab/method_compare}
  \end{minipage}

  \caption{\textbf{Comparison of video camera redirection paradigms.}
    We compare $3$ representative approaches for camera redirection from monocular video.
    \textbf{(A) Implicit Control:} Camera motion is specified via learned embeddings. Such implicit representations provide only soft controllability: text cannot precisely describe complex trajectories, and learned conditions often fail to follow the intended path (e.g., the ``turn to back'' instruction). Moreover, training requires paired data, which is expensive and scarce.
    \textbf{(B) Explicit Warping:} These methods warp observed pixels to target viewpoints using estimated depth. However, occluded regions remain unfilled, producing visible holes (red circles show ``Half Camel'').
    \textbf{(C) Ours:} We reconstruct a foreground-complete 4D proxy that recovers both visible and occluded foreground surfaces, then render it from target viewpoints as geometric guidance for video generation. This enables precise camera control with complete visibility from arbitrary viewpoints (green circle shows ``Complete Camel'').
    The table summarizes the key trade-offs. $\checkmark$ and $\times$ denote full and no support, respectively.}
    \Description{Comparison of video camera redirection paradigms with schematic diagrams of implicit control, explicit warping, and our foreground-complete approach, alongside a table summarizing the key trade-offs in camera controllability and visibility completeness.}
  \label{fig:combined_intro}
\end{figure}

Camera redirection seeks to endow machines with the ability to synthesize novel video sequences from a source video along a user-specified camera trajectory~\cite{he2024cameractrl,hou2024training,yu2024viewcrafter,bai2025recammaster}. For example, consider the video interview footage in Fig.~\ref{fig:teaser}. Although recorded from a fixed viewpoint, we can easily imagine the same scene from the side, back, or top-down, by mentally reconstructing a 3D world and replaying its dynamics from a virtual perspective. This capability is central to applications such as Autonomous Driving~\cite{cao2025pseudo,zhou2025predicting,zhou2023matters}, AR/VR~\cite{azuma1997survey}, and cinematic effects like bullet-time replay from limited camera setups~\cite{gao2021dynamic,tian2023mononerf,zhang2025stable,wang2025bullettime}. Today, in industry, such effects rely on costly multi-camera rigs and multi-view reconstruction techniques~\cite{mildenhall2021nerf,kerbl20233d,tancik2023nerfstudio,jiang2025anysplat,charatan2024pixelsplat,chen2024mvsplat}; enabling free-viewpoint camera redirection from a single video would democratize these experiences for everyday capture.

However, camera redirection from a single monocular video is inherently ill-posed. A monocular video captures only a narrow spatio-temporal view of a dynamic 3D scene, providing highly partial observations of the 4D world. The key challenge is to recover a complete and coherent plenoptic representation from this limited input, with consistent geometry and coherent motion. While prior methods can interpolate nearby novel views by reconstructing observed surfaces~\cite{mildenhall2021nerf,kerbl20233d,tian2023mononerf}, they break down under large-angle viewpoint changes far from the original camera trajectory. In these regimes, the absence of visual grounding introduces severe geometric ambiguity, fundamentally limiting replay from arbitrary viewpoints.

To tackle the challenge, recent studies~\cite{yu2024viewcrafter,hou2024training,li2025realcam,bai2025recammaster,luo2025camclonemaster,hu2025ex,yu2025trajectorycrafter,ren2025gen3c,zhang2025recapture,huang2025spacetimepilot,zheng2026versecrafter,gu2025diffusion,you2024nvs} try to explore the generative power of video foundation models for camera redirection.
As shown in Fig.~\ref{fig:combined_intro}, these methods generally follow two paradigms: implicit control and explicit warping.
Implicit control methods~\cite{bai2025recammaster,hou2024training,he2024cameractrl} encode camera trajectories as learned embeddings~\cite{bai2025recammaster,hou2024training,he2024cameractrl}  or text prompts~\cite{wang2025wan} and inject them into video diffusion models; however, such implicit representations offer only soft controllability and require expensive 4D training data.
Explicit warping methods~\cite{yu2025trajectorycrafter,hu2025ex,ren2025gen3c,seo2024genwarp} instead estimate depth and warp observed pixels to target viewpoints, but since only visible surfaces are available, occluded regions produce unfilled holes that must be hallucinated by downstream generators~\cite{wang2025wan,yang2024cogvideox}.
Neither paradigm achieves both precise camera control and complete visibility of unseen regions, both essential for faithful large-angle redirection.

In this paper, we propose FreeOrbit4D, an effective training-free framework for arbitrary camera redirection from a single monocular video via foreground-complete 4D reconstruction.
Our key idea is to explicitly reason about and recover a complete 4D geometry from partial monocular observations, and to use this geometry as structural guidance for generative rendering at novel viewpoints.
By completing unseen geometry, our method resolves the ambiguity that causes artifacts in prior work, enabling consistent synthesis even when the target trajectory deviates significantly from the original viewpoints.

Specifically, we observe that reconstructing dynamic scenes from monocular video and completing occluded object geometry are fundamentally different tasks: the former requires temporally consistent scene-level reasoning, while the latter demands multi-view understanding of object shape.
This motivates handling them in distinct representation spaces: we unproject the source video into a global scene space with geometrically incomplete foreground, and complete the foreground geometry via multi-view synthesis in canonical object space.
While this separation yields tractable sub-tasks addressable by existing techniques~\cite{zhou2025page,xie2024sv4d,wang2025vggt}, the key challenge lies in unifying their outputs into a coherent whole. Our correspondence-aware alignment addresses this by bridging canonical and global spaces through dense pixel-synchronized 3D–3D correspondences, producing a unified foreground-complete 4D proxy without task-specific training.
To interface this proxy with a video generator, we distill it into view-dependent depth maps as conditioning~\cite{wang2025wan}. While depth is not full geometry, it compactly captures the metric layout and visibility cues that enforce cross-view and temporal consistency. This conditioning enables faithful large-angle view synthesis and paves the way for potential 4D data generation.

Our contributions are summarized as follows:
\begin{itemize}
    \item We present a training-free method that achieves foreground-complete 4D reconstruction from monocular video by combining global scene lifting with object geometry completion through dense 3D correspondences.
    \item Building on this reconstruction, we propose a camera redirection framework tailored for large-angle view synthesis, enabling expansive camera motions with strong geometric consistency across time and viewpoints.
    \item Extensive experiments, including a user study, validate state-of-the-art performance under challenging large-angle trajectories. The explicit 4D representation further supports downstream applications by providing consistent geometry for video editing, editable structure for 4D manipulation, and dense annotations for potential 4D data generation.
\end{itemize}

%% file: tab/method_compare.tex
\resizebox{\linewidth}{!}{
\begin{tabular}{l | c c}
\toprule
\textbf{Method} &
\textbf{Camera Controllability} &
\textbf{Visibility Completeness} \\
\midrule

(A) Implicit Control &
$\times$\ \ Soft &
$\checkmark$\ \ Generative \\

(B) Explicit Warping &
$\checkmark$\ \ Warp-based &
$\times$\ \ Occlusion \\
\midrule

\textbf{(C) Ours} &
$\checkmark$\ \ Proxy-guided &
$\checkmark$\ \ FG-Complete \\
\bottomrule
\end{tabular}
}

%% file: sec/2_related_work.tex
\section{Related Work}
\label{subsec:overview}

\begin{figure*}[!t]
  \centering
    \includegraphics[width=\linewidth]{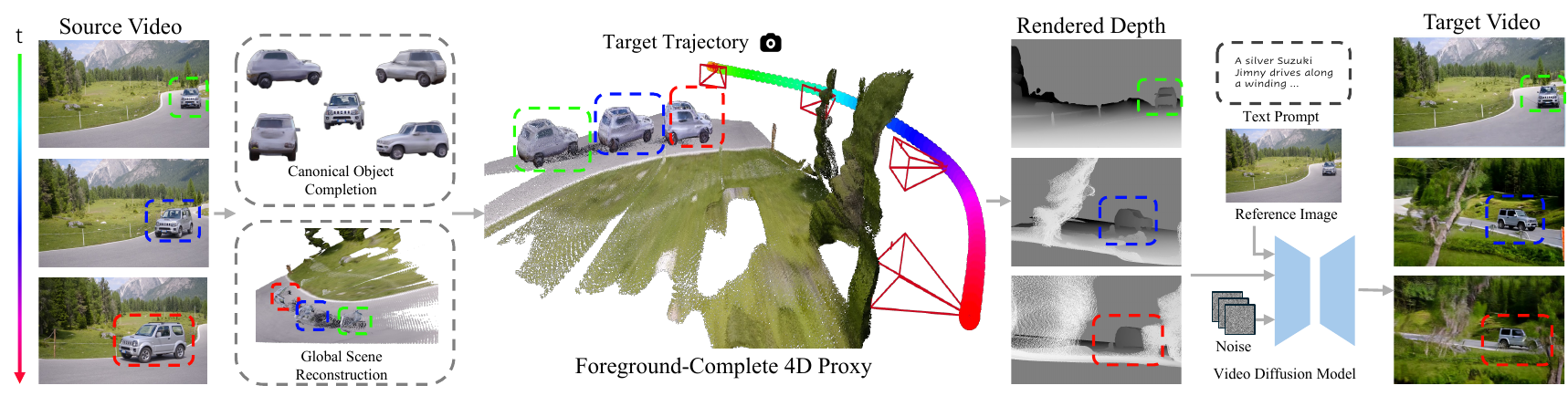}
    \caption{\textbf{Overview of FreeOrbit4D.}
    Our framework redirects a monocular video to a target camera trajectory via a foreground-complete 4D proxy. This proxy is constructed through two branches: \textit{Global Scene Reconstruction} recovers background and partial foreground in global space, while \textit{Canonical Object Completion} reconstructs complete foreground geometry via multi-view synthesis. After alignment, we render depth maps from the unified 4D proxy to condition a video diffusion model, along with a reference image and text prompt, enabling faithful novel-view synthesis under large viewpoint changes.}
    \Description{A pipeline diagram showing the FreeOrbit4D framework. The process is split into three main stages: First, Canonical Object Completion generates multi-view images to reconstruct complete foreground geometry. Second, Global Scene Reconstruction unprojects the source video to recover the background and partial foreground. Third, these are aligned and merged into a 4D proxy. Finally, depth scaffolds from the proxy guide a video diffusion model to produce the redirected target video.}
  \label{fig:pipeline}
\end{figure*}

\paragraph{Camera-controlled video generation}
With the rapid development of video foundation models~\cite{kong2024hunyuanvideo,blattmann2023stable}, research has progressed from pure text-to-video generation toward controllable video synthesis with multiple conditioning signals~\cite{zhang2023adding,bahmani2024vd3d,wang2024motionctrl}. Among these directions, camera-controlled video generation has attracted increasing attention due to its relevance to downstream video creation applications. Early approaches encode camera intrinsics and extrinsics as control signals~\cite{wang2024motionctrl,van2024generative}, but subsequent studies show that such low-dimensional parameters are difficult to exploit effectively in high-dimensional attention-based diffusion models. This has motivated alternative camera representations, including Plücker coordinates~\cite{he2024cameractrl,zhang2025spatialcrafter,fan2025omniview}, PRope~\cite{li2025cameras,huang2025voyager}, and per-token trajectory attention~\cite{xiao2024trajectory}, together with diverse control injection mechanisms~\cite{wang2024motionctrl,he2024cameractrl,huang2025voyager,bai2025recammaster,luo2025camclonemaster} and decoupled time--camera conditioning~\cite{wang2025bullettime}. This paradigm has been further extended to user-provided source videos: ReCapture~\cite{zhang2025recapture}, SpaceTimePilot~\cite{huang2025spacetimepilot}, and CAT4D~\cite{wu2025cat4d} re-render an input video along new camera trajectories via per-video fine-tuning or learned camera-motion disentanglement.
Despite promising results enabled by powerful diffusion priors, methods in this family often struggle to faithfully follow prescribed camera trajectories (Fig.~\ref{fig:pipeline}), especially under large-angle redirection, due to the absence of explicit, dense 4D geometric conditioning. In contrast, our training-free approach achieves precise camera redirection by reconstructing and completing 4D scene geometry.

\paragraph{Reconstruction-grounded 4D generation}
A second line of work uses explicit 3D or 4D geometry to ground camera-redirected video generation. Methods either inject lightweight cues such as rendered geometry buffers~\cite{gu2025diffusion,zheng2026versecrafter} or sparse 3D point tracks~\cite{lee2025generative} into video diffusion, or reconstruct a full 4D proxy of the scene from the source video~\cite{li2025realcam,hou2024training,tian2025voyaging,yu2025trajectorycrafter,cao2024motion2vecsets,tang2026motion2vecsets,yu2024viewcrafter,zhang2025joint,song2025worldforge,liu2025light,chen2025cognvs}. Pipelines that reconstruct a full 4D proxy rely on varying reconstruction backbones---3D representations such as NeRF~\cite{mildenhall2021nerf,tian2023mononerf,tian2024semantic} and Gaussian Splatting~\cite{kerbl20233d}, or 4D representations including motion scaffolds and dynamic Gaussians~\cite{wang2025shape,lei2025mosca,liao2025pad3r,luo2025instant4d,he2025restage4d}, video-native geometric estimators~\cite{zhang2024monst3r,li2025megasam,huang2025vipe,wang2025pi,sucar2026v,zhu2026motioncrafter,yao2025uni4d}, and feed-forward 4D predictors~\cite{jiang2025geo4d,lin2025movies}. Partial novel views are then rendered from the reconstructed geometry along the target trajectory, and the resulting unfilled holes are hallucinated by a downstream video diffusion model. However, the reconstructed geometry captures only visible surfaces from the source video, leaving the occluded foreground surfaces unrecovered in 3D. Under large-angle camera redirection, these unrecovered regions dominate the target view, and pixel-level hallucination without 3D support cannot synthesize them consistently, producing incomplete foreground and visible holes (Fig.~\ref{fig:pipeline}). A complementary thread of 4D generative models instead synthesizes 4D content end-to-end. Object-only variants~\cite{ren2023dreamgaussian4d,zeng2024stag4d,chen2026motion,li2024vivid} are confined to a single dynamic object, while scene-level variants, including multi-view video diffusion~\cite{lu2025see4d,mi2025one4d,li20244k4dgen,yuan2025seeu}, native 4D synthesis~\cite{li2025ss4d,yang2026neoverse}, and action-conditioned scene generation~\cite{li2025wonderplay,liu2026realwonder,zhan2026perpetualwonder}, do not ingest a user-provided source video and therefore cannot redirect existing footage along a prescribed camera trajectory. In contrast, our method reconstructs a foreground-complete 4D proxy via correspondence-aware alignment, recovering both visible and occluded foreground surfaces to enable precise, geometry-grounded redirection of user-provided monocular videos.

%% file: sec/3_method.tex
\section{Method}

In this work, we study the camera redirection task: Given a monocular source video $\mathcal{V}^{src}$ and a target camera trajectory, our goal is to synthesize a visually faithful and temporally consistent target video $\mathcal{V}^{tgt}$ that depicts the same dynamic scene as $\mathcal{V}^{src}$ but viewed from the target trajectory.

Fig.~\ref{fig:pipeline} shows our three-stage framework. First, we reconstruct a static background and geometrically incomplete foreground in global scene space, and complete foreground geometry in canonical object space (Sec.~\ref{subsec:reconstruction}).
Second, we align the canonical foreground to the global scene via dense pixel-synchronized 3D--3D correspondences, yielding a unified foreground-complete 4D proxy (Sec.~\ref{subsec:alignment}), from which we render depth scaffolds along the target trajectory and synthesize the output with a video diffusion model (Sec.~\ref{subsec:synthesis}).

\begin{figure*}[!t]
  \centering
   \includegraphics[width=\linewidth]{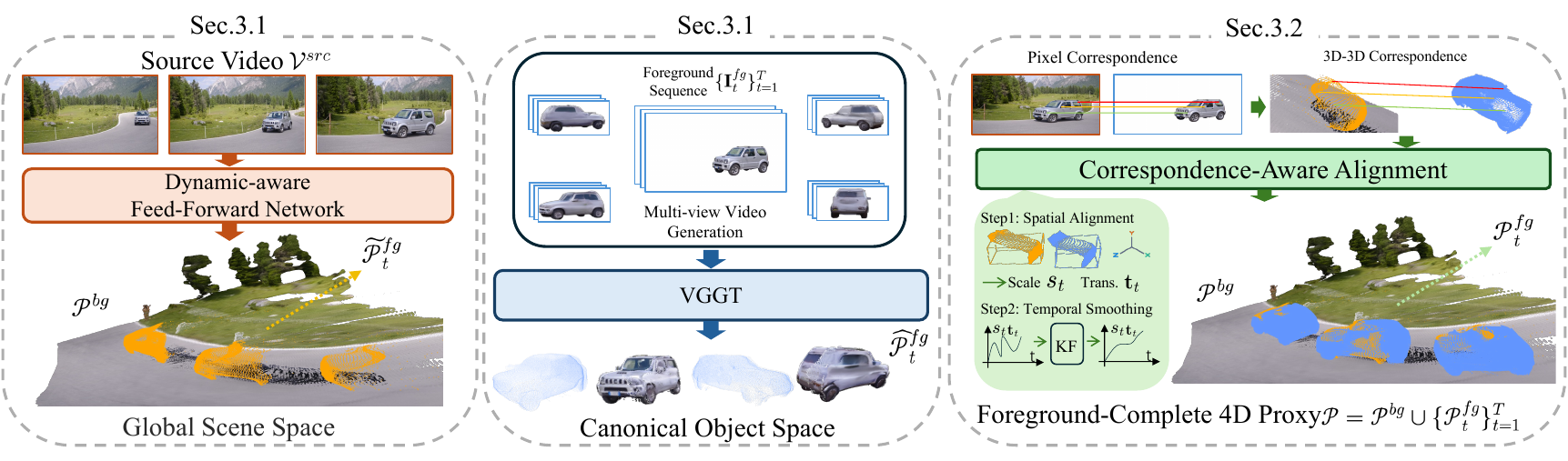}
    \caption{\textbf{Decoupled 4D reconstruction and alignment pipeline.}
    \textbf{Left (Sec.~\ref{subsec:reconstruction}):} A dynamic-aware feed-forward network lifts the source video $\mathcal{V}^{src}$ into \textit{global scene space}, producing the static background $\mathcal{P}^{bg}$ and geometrically incomplete foreground $\widetilde{\mathcal{P}}_t^{fg}$ (orange).
    \textbf{Middle (Sec.~\ref{subsec:reconstruction}):} The foreground sequence $\{\mathbf{I}_t^{fg}\}_{t=1}^T$ is fed into an object-centric video diffusion model to synthesize multi-view images, from which VGGT reconstructs complete foreground geometry $\widehat{\mathcal{P}}_t^{fg}$ in \textit{canonical object space}.
    \textbf{Right (Sec.~\ref{subsec:alignment}):} Dense 3D-3D correspondences derived from pixel-synchronized point maps enable \textit{correspondence-aware alignment}: per-frame spatial alignment estimates scale $s_t$ and translation $\mathbf{t}_t$, followed by temporal smoothing via Kalman filtering. The result is the unified foreground-complete 4D proxy $\mathcal{P} = \mathcal{P}^{bg} \cup \{\mathcal{P}_t^{fg}\}_{t=1}^T$.}
    \Description{A detailed technical diagram showing the decoupled reconstruction and alignment pipeline. On the left, a source video is processed to separate the static background and the incomplete foreground in global scene space. The middle section shows the foreground sequence being processed through an object-centric diffusion model and VGGT to create complete geometry in canonical object space. The right section illustrates the alignment process where dense correspondences, scale estimation, and Kalman filtering are used to merge the canonical foreground back into the global scene, forming a unified 4D proxy.}
  \label{fig:method}
\end{figure*}

\subsection{Decoupled 4D Reconstruction}
\label{subsec:reconstruction}
Since the static scene and the moving object exhibit fundamentally different geometric and motion characteristics, it is natural to decouple their reconstructions. We therefore reconstruct geometry in two complementary coordinate spaces. In \textbf{global scene space}, we recover the static background point cloud $\mathcal{P}^{bg}$ and partial foreground point clouds $\widetilde{\mathcal{P}}_t^{fg}$, which are lifted from the source frames and thus capture only visible surfaces. In \textbf{canonical object space}, we reconstruct complete foreground point clouds $\widehat{\mathcal{P}}_{t}^{fg}$ that capture the full 3D structure but lack alignment to the global scene. Throughout, we use tilde ( $\widetilde{\cdot}$ ) to denote geometrically incomplete representations in global scene space, and hat ( $\widehat{\cdot}$ ) to denote geometrically complete but unaligned representations in canonical object space. These are unified in Sec.~\ref{subsec:alignment}.

\paragraph{Global Scene Reconstruction}
VGGT~\cite{wang2025vggt} is a feed-forward model that predicts 3D point maps from images, but assumes a static scene. To handle dynamic videos, we adopt its temporally-aware extension~\cite{zhou2025page}, which processes the entire monocular sequence $\mathcal{V}^{src}$ and predicts temporally consistent point maps $\widetilde{\mathbf{P}}_t \in \mathbb{R}^{H \times W \times 3}$ for each frame, all registered in a unified global coordinate system.

We then use semantic masks $\mathbf{M}_{t}$ from SAM2~\cite{ravi2024sam} to separate background and foreground, initiated by a user-provided point prompt on the first frame, or fully automated via detector-based prompting~\cite{ren2024grounded}. The foreground-background decomposition is:
\begin{equation}
    \mathcal{P}^{bg} = \bigcup_{t=1}^{T} \{ \widetilde{\mathbf{P}}_t(\mathbf{u}) \mid \mathbf{M}_t(\mathbf{u}) = 0 \}, \quad
    \widetilde{\mathcal{P}}_t^{fg} = \{ \widetilde{\mathbf{P}}_t(\mathbf{u}) \mid \mathbf{M}_t(\mathbf{u}) = 1 \}.
\end{equation}
Here, $\mathbf{u}$ denotes pixel coordinates, and $\widetilde{\mathcal{P}}_t^{fg}$ indicates partial foreground point clouds in global space, as only visible surfaces are captured from the source view (orange in Fig.~\ref{fig:method}).

\paragraph{Canonical Object Completion}
To resolve the inherent single-view ambiguity in foreground geometry, we leverage a multi-view video diffusion model~\cite{yao2025sv4d} to synthesize novel views. Since this model is object-centric and requires isolated object inputs, we first extract the foreground from each frame $\mathbf{I}_t$ of the source video $\mathcal{V}^{src}$ using the semantic masks, yielding the masked foreground sequence $\{\mathbf{I}_{t}^{fg}\}_{t=1}^T$. The model then synthesizes four temporally synchronized novel-view videos at $90^\circ$ azimuthal intervals, denoted as $\{\mathbf{I}_t^{(k)}\}_{t=1}^T$ for $k = 1, \dots, 4$.

We then use VGGT~\cite{wang2025vggt} to reconstruct multi-view point maps from all five views (source + four synthesized) per frame:
\begin{equation}
    \widehat{\mathbf{P}}_{t}^{fg} = \Phi_{\text{VGGT}}\left(\mathbf{I}_{t}^{fg}, \mathbf{I}_t^{(1)}, \dots, \mathbf{I}_t^{(4)}\right), \quad \widehat{\mathbf{P}}_{t}^{fg} \in \mathbb{R}^{5 \times H \times W \times 3}
\end{equation}
where the source view $\mathbf{I}_{t}^{fg}$ serves as the reference. $\widehat{\mathbf{P}}_{t}^{fg}$ resides in canonical object space and requires alignment to the global scene.

Since the synthesized multi-view images contain white backgrounds, the raw point maps include redundant points. We filter them using binary masks: the SAM2 mask $\mathbf{M}_t$ for the source view, and color thresholding masks $\{\mathbf{M}_t^{(k)}\}_{k=1}^4$ for the novel views. Let $\widehat{\mathbf{P}}_t$ and $\widehat{\mathbf{P}}_t^{(k)}$ denote the point maps for the source and $k$-th novel view, respectively. The complete canonical foreground point cloud is:
\begin{equation}
    \widehat{\mathcal{P}}_{t}^{fg} = \{ \widehat{\mathbf{P}}_t(\mathbf{u}) \mid \mathbf{M}_t
    (\mathbf{u}) = 1 \} \cup \bigcup_{k=1}^{4} \{ \widehat{\mathbf{P}}_t^{(k)}(\mathbf{u}) \mid \mathbf{M}_t^{(k)}(\mathbf{u}) = 1 \}
    \label{eq:foreground-complete}
\end{equation}

\begin{figure*}[!t]
  \centering
    \includegraphics[width=0.9\linewidth]{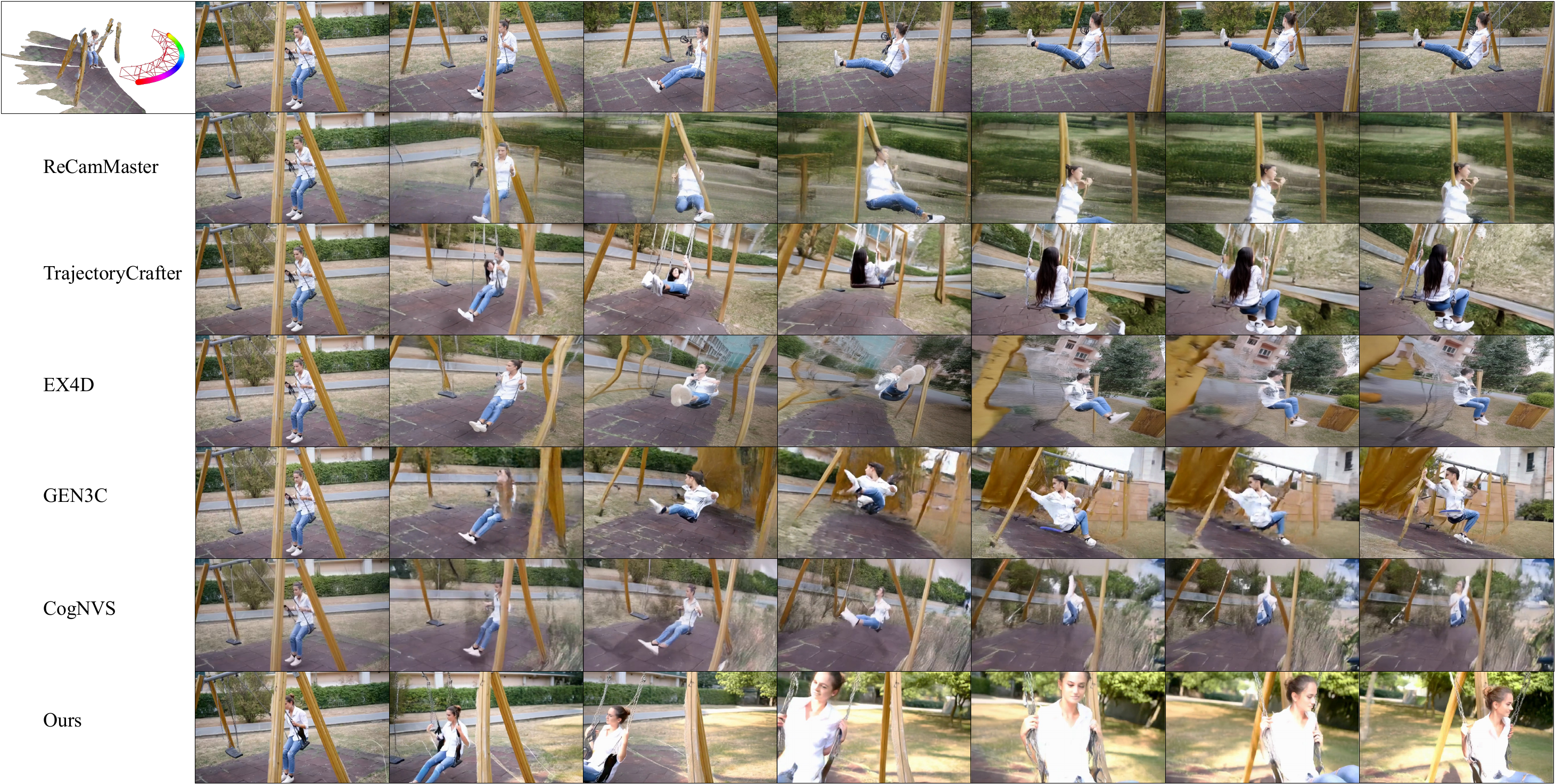}
    \caption{\textbf{Qualitative comparison on the ``Swing'' sequence.} The top-left shows the target camera trajectory orbiting around the scene. This challenging scenario involves rapid foreground motion and thin structures (swing ropes). ReCamMaster~\cite{bai2025recammaster} and EX-4D~\cite{hu2025ex} fail to preserve human body structure, producing blurred limbs and ghosting artifacts. TrajectoryCrafter~\cite{yu2025trajectorycrafter}, GEN3C~\cite{ren2025gen3c}, and CogNVS~\cite{chen2025cognvs} exhibit noticeable geometric distortions and semantic drift as the camera rotates. In contrast, our method maintains sharp details and stable geometry throughout the sequence, benefiting from the foreground-complete 4D proxy. More visualizations are provided in the supplementary material.}
    \Description{A qualitative comparison on the 'Swing' sequence. The image layout features the target camera trajectory in the top-left, followed by a grid of video frame comparisons between different methods. Baselines including ReCamMaster, EX-4D, TrajectoryCrafter, and GEN3C show visible artifacts such as blurred limbs, ghosting, and geometric distortions as the camera orbits. The bottom row displays our method's results, which maintain sharp human body structures and temporal consistency throughout the large-angle rotation.}
    \label{fig:swing}
\end{figure*}

\input{tab/vbench}

\subsection{Correspondence-Aware Alignment}
\label{subsec:alignment}
We now have partial foreground point clouds $\widetilde{\mathcal{P}}_{t}^{fg}$ in global scene space, and complete canonical foreground point clouds $\widehat{\mathcal{P}}_{t}^{fg}$ in canonical object space. Our goal is to align $\widehat{\mathcal{P}}_{t}^{fg}$ to global scene space, yielding the aligned foreground point clouds $\mathcal{P}_{t}^{fg}$, which together with the static background $\mathcal{P}^{bg}$ form the unified foreground-complete 4D proxy $\mathcal{P} = \mathcal{P}^{bg} \cup \{\mathcal{P}_{t}^{fg}\}_{t=1}^{T}$.

\paragraph{Per-frame Spatial Alignment}
We estimate a per-frame transformation $\mathcal{T}_t$ such that $\mathcal{P}_{t}^{fg} = \mathcal{T}_t(\widehat{\mathcal{P}}_{t}^{fg})$, using $\widetilde{\mathcal{P}}_{t}^{fg}$ as the spatial anchor.
A key observation is that both $\widetilde{\mathbf{P}}_t$ and $\widehat{\mathbf{P}}_t$ originate from the same source image $\mathbf{I}_t$, so points at the same pixel $\mathbf{u}$ correspond to the same surface point. This yields dense 3D--3D correspondences:
\begin{equation}
\mathcal{C}_t = \{(\widehat{\mathbf{P}}_t(\mathbf{u}), \widetilde{\mathbf{P}}_t(\mathbf{u})) \mid \mathbf{M}_t(\mathbf{u}) = 1\},
\end{equation}
which we filter by confidence and outlier removal.

Given $\mathcal{C}_t$, a natural approach would be to minimize point-wise distances. However, monocular lifting cannot determine absolute depth scale, which may vary across frames, causing per-point inconsistencies in $\widetilde{\mathcal{P}}_{t}^{fg}$. While point-wise fitting would propagate these errors, the overall spatial extent and location of $\widetilde{\mathcal{P}}_{t}^{fg}$ within the scene remain reliable. In contrast, $\widehat{\mathcal{P}}_{t}^{fg}$ provides accurate object geometry from multi-view reconstruction but lacks scene-level positioning. We therefore use $\widetilde{\mathcal{P}}_{t}^{fg}$ only to determine the global placement of the object (position and scale), while preserving the geometry of $\widehat{\mathcal{P}}_{t}^{fg}$. Specifically, we parameterize $\mathcal{T}_t$ as:
$
s_t \, \mathbf{x} + \mathbf{t}_t,
$
where $(s_t, \mathbf{t}_t)$ are estimated from $\mathcal{C}_t$ such that the transformed geometry matches the scale and location of $\widetilde{\mathcal{P}}_{t}^{fg}$. Rotation is fixed since both reconstructions use the source view as coordinate reference.

\paragraph{Temporal Smoothing}
The geometrically incomplete foreground $\widetilde{\mathcal{P}}_{t}^{fg}$ may exhibit frame-to-frame depth inconsistency due to the inherent ambiguity of monocular lifting. Since we use it to anchor global placement, this inconsistency propagates into the aligned foreground trajectory. To compensate, we smooth the centroid trajectory of aligned point clouds using a bidirectional Kalman filter with a constant-velocity motion model, with stronger regularization along depth. The per-frame scale is preserved for geometric fidelity, yielding a temporally coherent sequence $\{\mathcal{P}_{t}^{fg}\}_{t=1}^{T}$.

\subsection{Geometry-conditioned Video Synthesis}
\label{subsec:synthesis}
Given the foreground-complete 4D proxy $\mathcal{P} = \mathcal{P}^{bg} \cup \{\mathcal{P}_{t}^{fg}\}_{t=1}^{T}$ and a target camera trajectory $\{\boldsymbol{\pi}_t^{tgt}\}_{t=1}^{T}$, we render depth scaffolds to guide video synthesis:
\begin{equation}
\mathcal{V}^{tgt} = \Phi_{\text{VDM}}\left(\mathbf{I}_1,\ \left\{\text{Render}\left(\mathcal{P},\ \boldsymbol{\pi}_t^{tgt}\right)\right\}_{t=1}^{T},\ \mathbf{c}\right),
\end{equation}
where $\Phi_{\text{VDM}}$ is a depth-conditioned video diffusion model~\cite{wang2025wan,jiang2025vace}, $\mathbf{I}_1$ is the first frame of $\mathcal{V}^{src}$ as appearance reference, and $\mathbf{c}$ is a text prompt. Depth maps form a compact yet informative geometric scaffold: they encode the 3D layout from the target viewpoint, allowing video diffusion to generate spatially coherent content that faithfully follows the prescribed camera motion.

%% file: tab/vbench.tex
\begin{table*}[t!]
\centering
\small
\caption{\textbf{Quantitative comparison and User Study.} We report VBench for perceptual video quality, DINO/CLIP-SIM for semantic similarity, FID-V/FVD-V for distributional fidelity, and user ratings (1--5 scale). \textbf{Bold}: best; \underline{underline}: second-best.}
\Description{A numerical results table comparing our method against five baselines (ReCamMaster, TrajectoryCrafter, EX-4D, GEN3C, CogNVS) across three metric groups: VBench perceptual quality (Subject Consistency, Background Consistency, Motion Smoothness, Overall Consistency, Aesthetic Quality, Imaging Quality), similarity and fidelity (DINO-SIM, CLIP-SIM, FID-V, FVD-V), and user study ratings (Overall, Motion, Stability on a 1 to 5 scale). Our method achieves the best scores on most metrics, with best values shown in bold and second-best values underlined.}
\setlength{\tabcolsep}{4pt}
\resizebox{\textwidth}{!}{%
\begin{tabular}{l | c c c c c c | c c c c | c c c}
\toprule
& \multicolumn{6}{c|}{VBench $\uparrow$} & \multicolumn{4}{c|}{Similarity \& Fidelity} & \multicolumn{3}{c}{User Study} \\
\cmidrule(lr){2-7} \cmidrule(lr){8-11} \cmidrule(lr){12-14}
\multirow{2}{*}{Method} & Subject & BG & Motion & Overall & Aesth. & Imaging & DINO-SIM & CLIP-SIM & FID-V & FVD-V & Overall & Motion & Stab. \\
& Consis. & Consis. & Smooth. & Consis. & Qual. & Qual. & ($\uparrow$) & ($\uparrow$) & ($\downarrow \times 10^2$) & ($\downarrow \times 10^3$) & ($\uparrow$) & ($\uparrow$) & ($\uparrow$) \\
\midrule
ReCamMaster       & \underline{0.84} & \underline{0.92} & \textbf{0.98} & 0.16 & 0.39 & 43 & 0.37 & 0.75 & 2.6 & 3.9 & 2.0 & 2.5 & 2.0 \\
TrajectoryCrafter & 0.80 & 0.91 & 0.94 & \underline{0.19} & \underline{0.47} & \underline{53} & \underline{0.47} & \underline{0.79} & \underline{2.0} & 3.6 & \underline{2.8} & 3.2 & \underline{2.9} \\
EX-4D             & 0.76 & 0.89 & 0.94 & 0.16 & 0.42 & 46 & 0.28 & 0.69 & 3.2 & 3.8 & 2.0 & 2.5 & 2.0 \\
GEN3C             & 0.79 & 0.88 & 0.95 & 0.18 & 0.42 & 49 & 0.43 & 0.75 & 2.3 & \textbf{3.3} & 2.4 & \underline{3.5} & 2.3 \\
CogNVS            & 0.82 & 0.91 & \underline{0.97} & 0.15 & 0.42 & \underline{53} & 0.34 & 0.78 & 3.0 & \underline{3.4} & 2.0 & 3.2 & 1.9 \\
\midrule
\textbf{Ours}     & \textbf{0.88} & \textbf{0.94} & 0.96 & \textbf{0.24} & \textbf{0.52} & \textbf{64} & \textbf{0.65} & \textbf{0.84} & \textbf{1.7} & 3.6 & \textbf{4.6} & \textbf{4.5} & \textbf{4.5} \\
\bottomrule
\end{tabular}%
}
\label{tab:quantitative}
\end{table*}

%% file: sec/4_experiments.tex
\section{Experiments and Applications}
\label{sec:experiments}

\begin{figure*}[!t]
  \centering
  \includegraphics[width=0.9\linewidth]{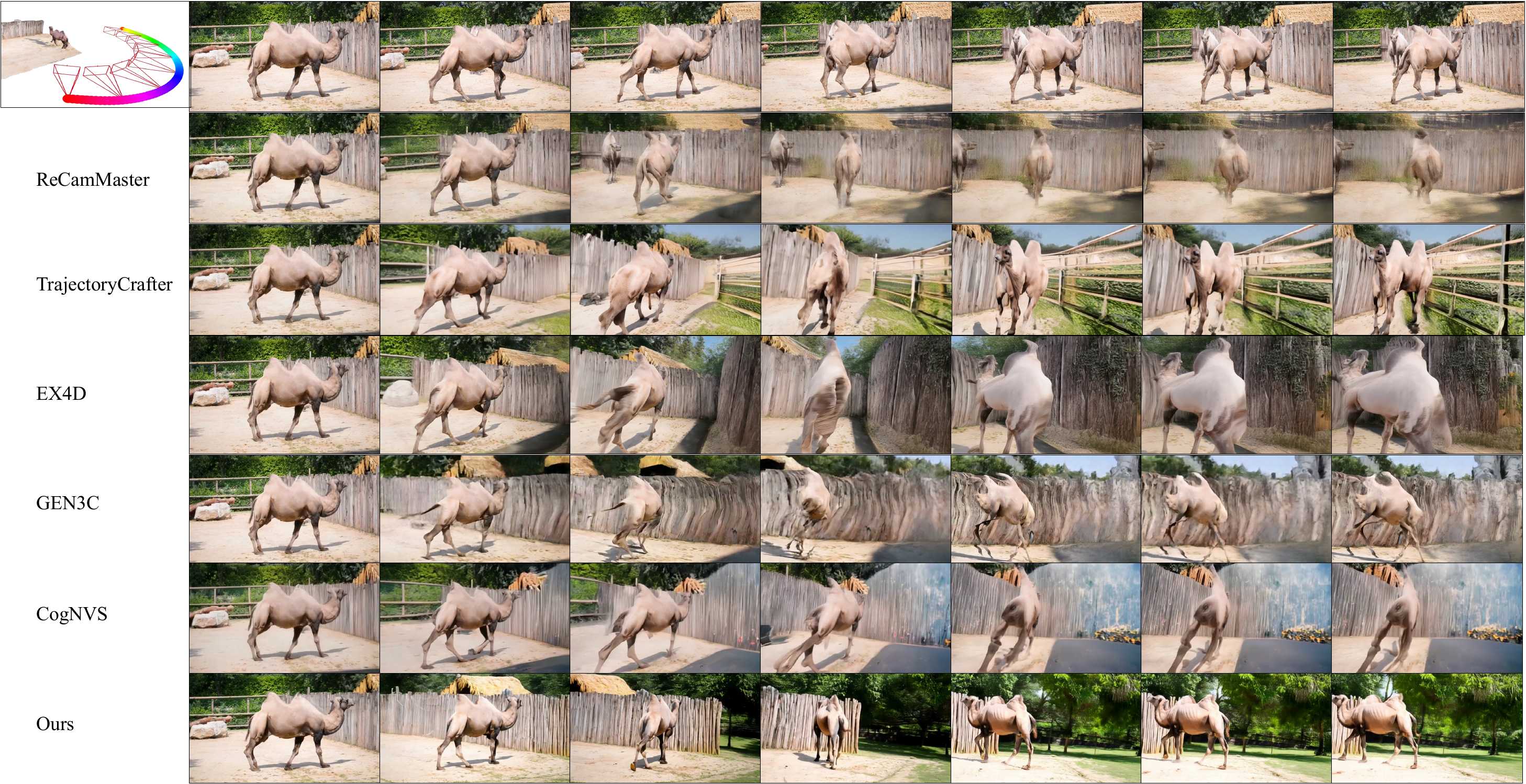}
  \includegraphics[width=0.9\linewidth]{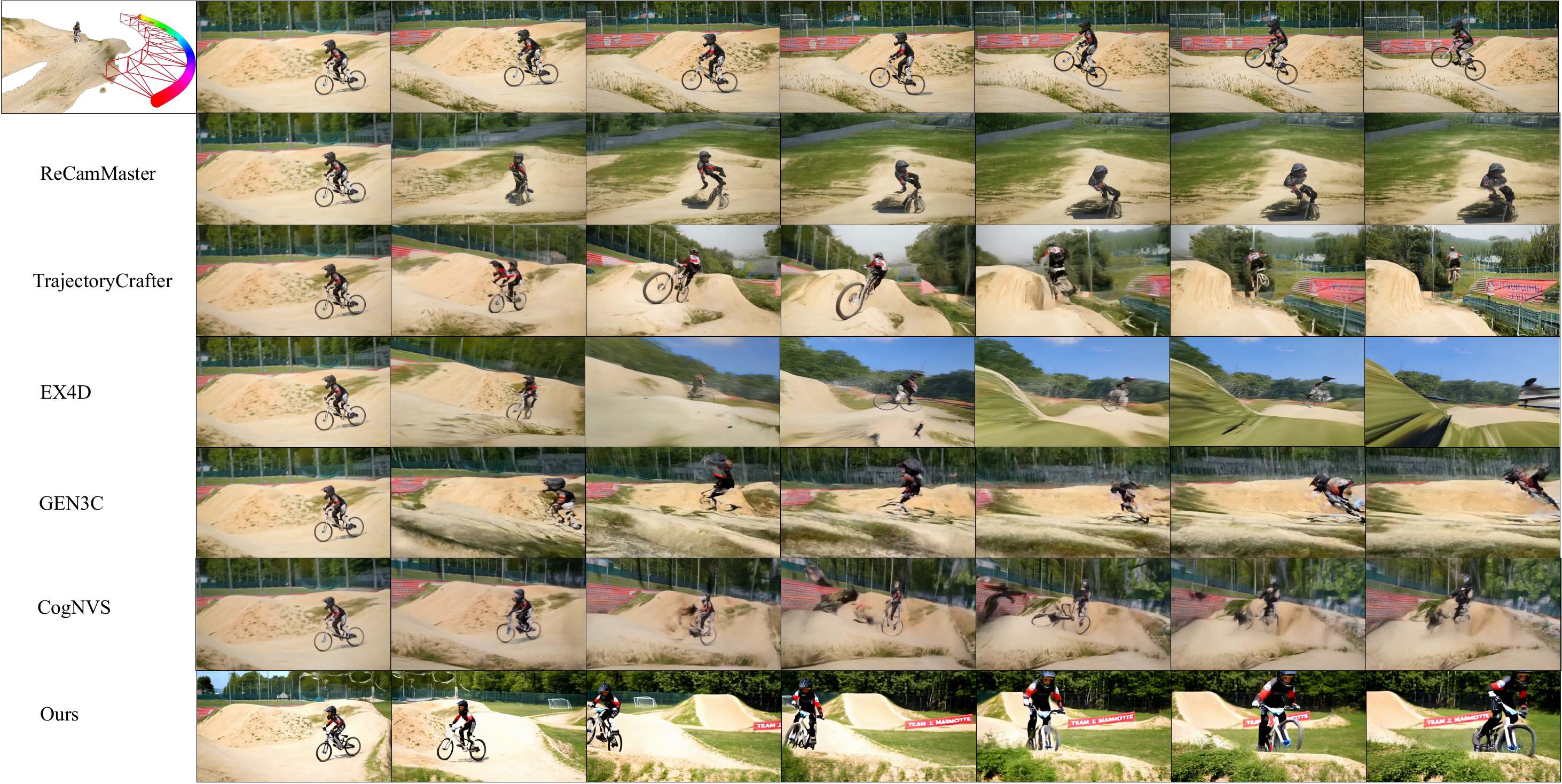}
  \caption{\textbf{Additional qualitative comparisons on ``Camel'' (top) and ``BMX'' (bottom).} Compared to baselines that suffer from geometric distortions, motion blur, and semantic drift under large viewpoint changes, our method produces consistently sharp textures and stable geometry across diverse scenes, demonstrating the effectiveness of our foreground-complete 4D proxy.}
  \Description{Qualitative comparison grid across multiple methods on the Camel and BMX sequences, showing that baselines exhibit distortion while our method maintains sharp textures and stable geometry.}
  \label{fig:qualitative_comparison}
\end{figure*}

\begin{figure*}[!t]
  \centering
  \includegraphics[width=0.9\linewidth]{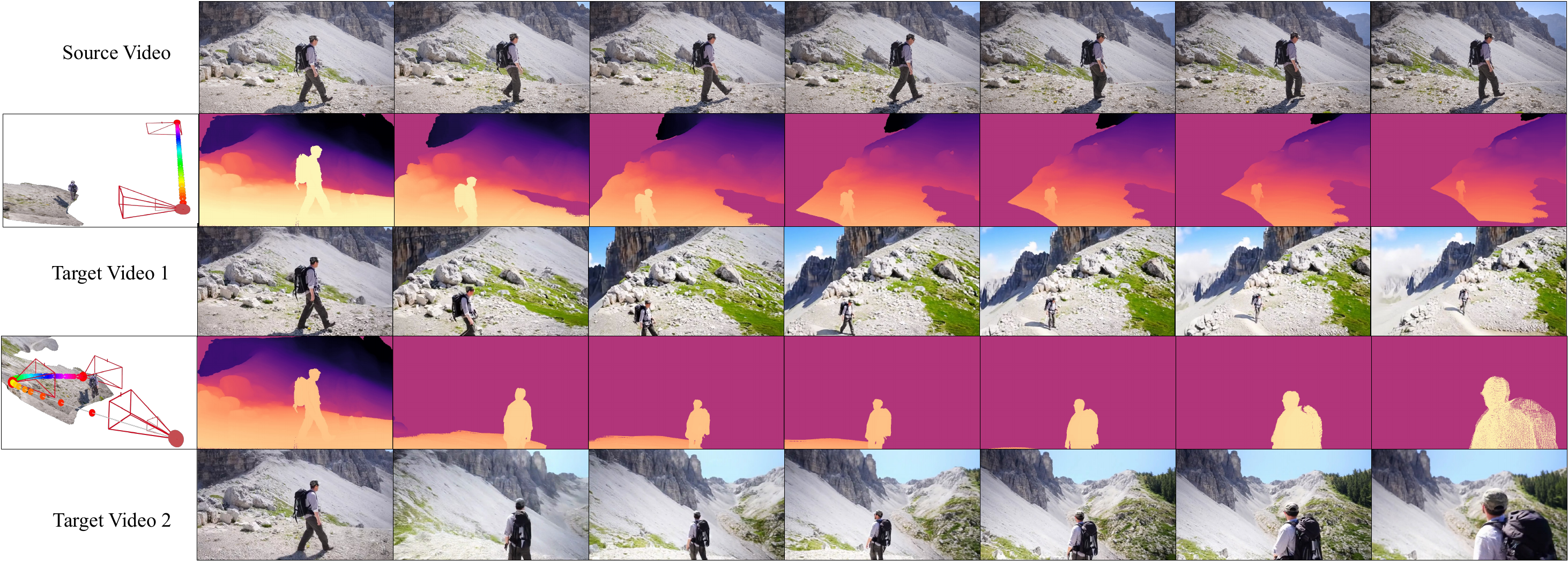}
  \includegraphics[width=0.9\linewidth]{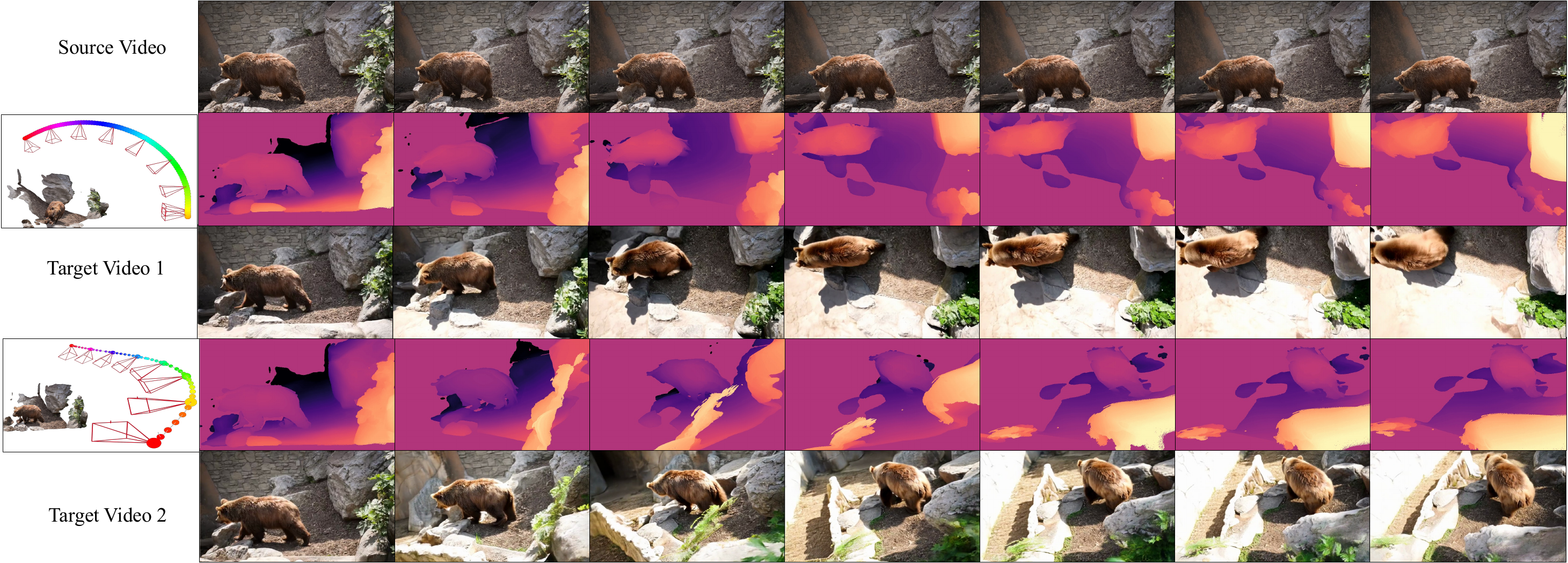}
  \includegraphics[width=0.9\linewidth]{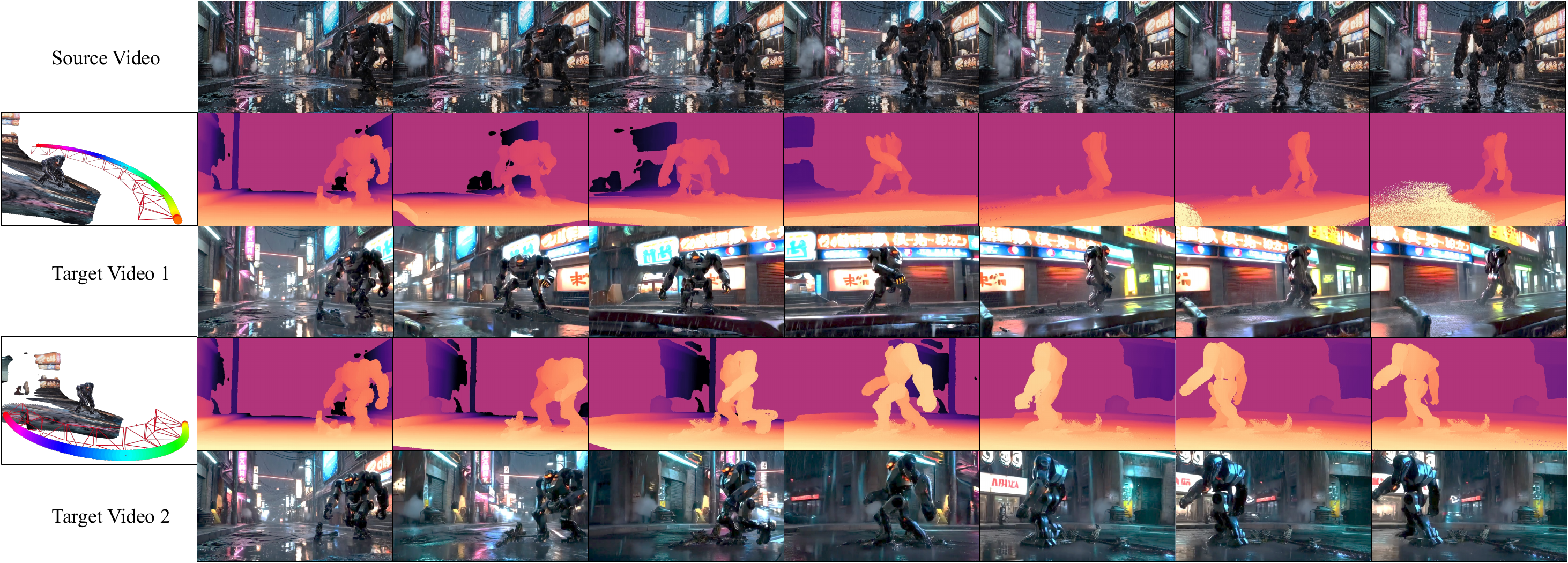}
  \caption{\textbf{Multi-trajectory video synthesis results on ``Hike'' (top), ``Bear'' (middle), and ``Robot'' (bottom).} We show that our method generates temporally and geometrically consistent videos along diverse target camera trajectories. By leveraging the foreground-complete 4D proxy, our approach effectively handles articulated motion, complex lighting, and thin structures across all sequences.}
  \Description{Multi-trajectory synthesis results on Hike, Bear, and Robot sequences. Each sub-figure shows consistent video frames and corresponding depth scaffolds rendered from different viewpoints.}
  \label{fig:multi_trajectory_results}
\end{figure*}

We evaluate FreeOrbit4D on diverse real-world and synthetic videos under challenging large-angle camera trajectories. After describing implementation details, datasets, and metrics, we compare against state-of-the-art methods (Sec.~\ref{subsec:comparison}), demonstrate practical applications (Sec.~\ref{subsec:applications}), and provide ablation studies (Sec.~\ref{sec:ablation}).

\paragraph{Implementation Details}
Our pipeline is training-free and relies on off-the-shelf pretrained models: PAGE-4D~\cite{zhou2025page} for global scene reconstruction, SAM2~\cite{ravi2024sam} for foreground segmentation, SV4D2.0~\cite{yao2025sv4d} for multi-view video synthesis, VGGT~\cite{wang2025vggt} for multi-view point-map reconstruction, and Wan2.2-VACE~\cite{wang2025wan,jiang2025vace} for depth-conditioned video synthesis.
All experiments use a single NVIDIA A40 GPU, processing each 45-frame clip at $832\times480$ in roughly 50~minutes end-to-end.

\paragraph{Evaluation Datasets}
We evaluate on a diverse set of real-world and synthetic videos.
For real-world data, we use sequences from DAVIS~\cite{perazzi2016benchmark}, which provides high-quality monocular videos with varied dynamic foregrounds, as well as publicly available online videos (e.g., Unitree robot demos, LeCun interview footage) that feature complex real-world motion and cluttered backgrounds.
For synthetic data, we include sequences generated by VEO~\cite{deepmind_veo_model_page} and Sora~\cite{openai_sora_creating_video_from_text}, covering diverse scenes and motion patterns.
To evaluate large-angle camera redirection, we define target trajectories with extreme yaw or pitch rotations (e.g., $120^\circ$ or $180^\circ$) from the initial viewpoint.

\paragraph{Evaluation Metrics}
We measure visual quality with FID-V~\cite{heusel2017gans} and FVD-V~\cite{unterthiner2019fvd}, semantic consistency with CLIP-SIM~\cite{radford2021learning} and DINO-SIM~\cite{oquab2023dinov2}, and perceptual video quality with VBench~\cite{huang2024vbench}. A user study further evaluates overall preference, camera accuracy, and temporal stability.

\subsection{Comparison with State-of-the-Art Methods}
\label{subsec:comparison}
\paragraph{Baselines} We compare our method with state-of-the-art camera-controlled video-to-video generation methods, including ReCamMaster~\cite{bai2025recammaster}, TrajectoryCrafter~\cite{yu2025trajectorycrafter}, EX-4D~\cite{hu2025ex}, GEN3C~\cite{ren2025gen3c}, and CogNVS~\cite{chen2025cognvs}.
ReCamMaster~\cite{bai2025recammaster} conditions a video generator on target camera poses for single-video re-rendering.
TrajectoryCrafter~\cite{yu2025trajectorycrafter} uses diffusion with point-cloud rendering guidance to follow a user-specified camera trajectory.
EX-4D~\cite{hu2025ex} leverages a depth-watertight mesh to improve extreme-viewpoint synthesis.
GEN3C~\cite{ren2025gen3c} maintains and renders a 3D point-cloud cache to enforce 3D-consistent camera control.
CogNVS~\cite{chen2025cognvs} completes disoccluded regions in rendered novel views via test-time fine-tuned video diffusion.

\paragraph{Qualitative Results} We present qualitative comparisons in Fig.~\ref{fig:swing}, Fig.~\ref{fig:qualitative_comparison}, and Fig.~\ref{fig:multi_trajectory_results}.
In the ``Swing'' scenario (Fig.~\ref{fig:swing}), which involves rapid foreground motion and thin structures (swing ropes), TrajectoryCrafter, GEN3C, and CogNVS~\cite{chen2025cognvs} exhibit geometric distortions and semantic drift as the camera rotates. ReCamMaster and EX-4D fail to preserve the structural integrity of the human body, producing blurred limbs or ghosting artifacts under large viewpoint changes.
In contrast, our method generates high-fidelity frames with sharp details and stable geometry by leveraging the foreground-complete 4D proxy to resolve single-view ambiguity.
Fig.~\ref{fig:qualitative_comparison} provides additional comparisons across diverse scenes, and Fig.~\ref{fig:multi_trajectory_results} showcases results under multiple user-specified trajectories, highlighting the flexibility of our trajectory control.

\paragraph{Quantitative Results}
Tab.~\ref{tab:quantitative} summarizes the quantitative comparison across automatic metrics and user ratings.
On VBench, our method ranks first in five out of six dimensions, achieving the highest subject consistency (0.88), background consistency (0.94), overall consistency (0.24), aesthetic quality (0.52), and imaging quality (64). Motion smoothness (0.96) ranks behind ReCamMaster (0.98) and CogNVS (0.97), both of which tend to produce over-smoothed results at the cost of geometric detail.
For semantic similarity and distributional fidelity, we achieve the best DINO-SIM (0.65), CLIP-SIM (0.84), and FID-V ($1.7 \times 10^2$), with competitive FVD-V ($3.6 \times 10^3$).
Despite these strong results, we observe that automatic metrics do not fully capture trajectory control quality. As shown in Fig.~\ref{fig:swing}, methods such as ReCamMaster and TrajectoryCrafter achieve reasonable scores while still deviating from the prescribed path or losing geometric coherence, motivating our user study.

\begin{figure}[!ht]
  \centering
  \includegraphics[width=\linewidth,trim={10 10 0 5},clip]{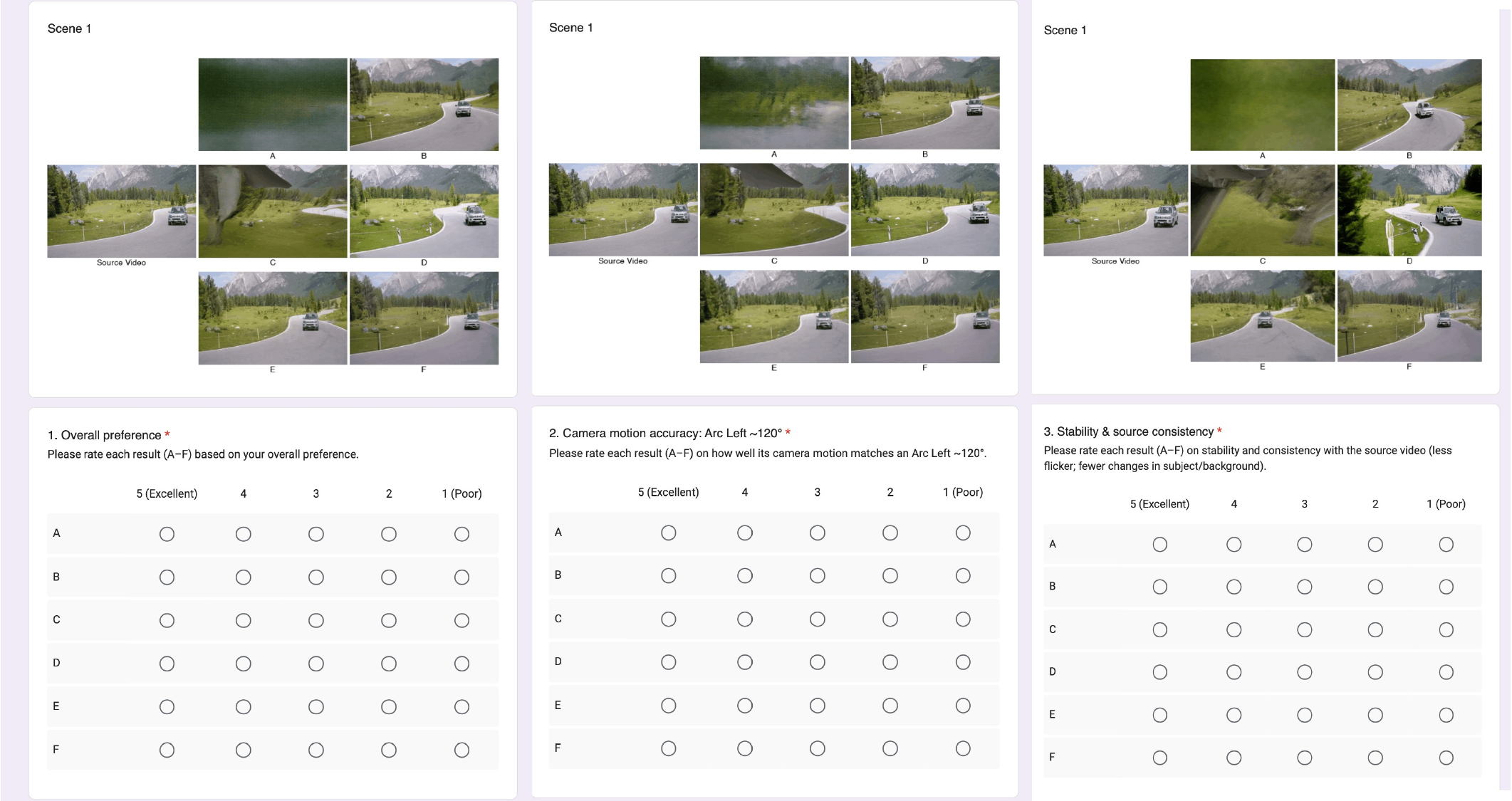}
  \caption{\textbf{User study interface.} We present participants with two anonymized videos (ours vs. baseline) and ask them to select the one with superior temporal stability and geometric fidelity.}
  \Description{Screenshot of the user study questionnaire interface showing two anonymized videos side by side with rating options.}
  \label{fig:user_study}
\end{figure}

\begin{figure}[!ht]
  \centering
    \includegraphics[width=\linewidth]{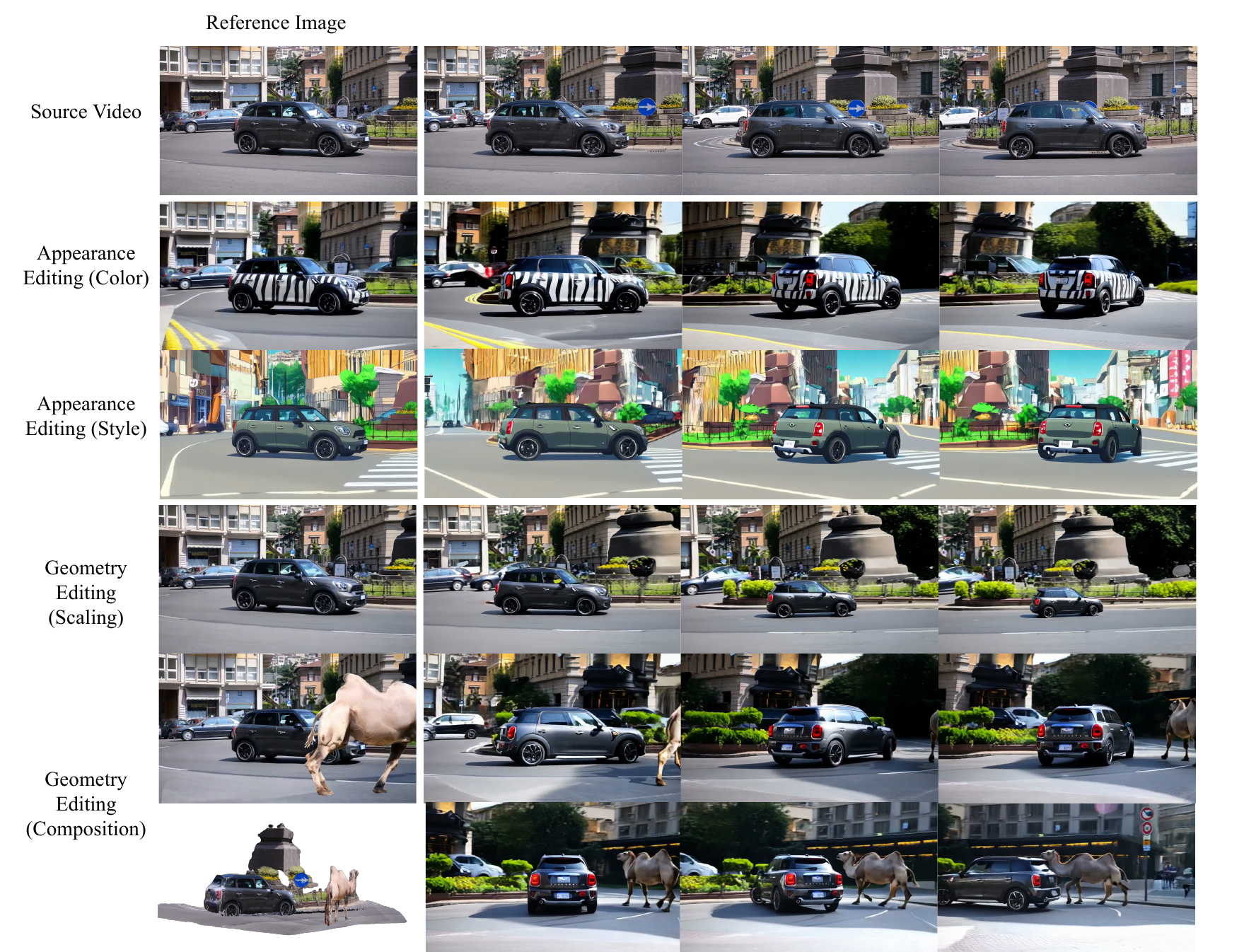}
    \caption{\textbf{Applications enabled by FreeOrbit4D.}
    Our explicit 4D representation enables various downstream applications.
    \textbf{Top:} Appearance editing---given a single edited reference frame (e.g., zebra pattern or anime style), our foreground-complete proxy propagates the edit consistently across all novel viewpoints.
    \textbf{Bottom:} Geometry editing---by directly manipulating the point cloud (scaling or compositing objects from different sources), we synthesize plausible redirected videos from the modified 4D geometry.}
    \Description{A grid of images demonstrating two main applications of FreeOrbit4D. The top half shows appearance propagation, where a single edited frame (black-and-white stripes and anime style) is consistently applied to a full redirected video sequence. The bottom half shows 4D geometry manipulation, illustrating direct edits to the point-based representation, such as scaling objects and composing a camel from another point cloud into the existing scene.}
  \label{fig:application}
\end{figure}

\paragraph{User Study} We conducted a user study with 20 participants across 10 diverse sequences. Participants rated results on a 1--5 scale along three axes (see the questionnaire interface in Fig.~\ref{fig:user_study}): (1) overall preference, (2) camera motion accuracy (adherence to the prescribed trajectory), and (3) temporal stability and source consistency (flicker and identity preservation). As shown in Tab.~\ref{tab:quantitative}, our method outperforms all baselines across all axes, with particularly large margins in motion accuracy (4.5 vs. 3.5 for the next best method). This validates that our foreground-complete 4D proxy provides precise camera control and structural stability that traditional metrics fail to capture, yielding an overall preference of 4.6.

\subsection{Applications}
\label{subsec:applications}

As shown in Fig.~\ref{fig:application}, FreeOrbit4D also enables several practical and interesting applications beyond camera redirection, thanks to our explicit and foreground-complete 4D representation.

\paragraph{Appearance Propagation}
Our framework propagates appearance edits from a single reference frame to the entire redirected video. We edit the reference frame with Qwen-Image-Edit~\cite{wu2025qwen}, applying color or style changes. The foreground-complete 4D proxy then provides consistent depth scaffolds, enabling faithful appearance transfer across all viewpoints.

\paragraph{4D Geometry Manipulation}
Our explicit point-based 4D representation enables direct geometric manipulation in space-time. By modifying the point cloud, such as scaling the foreground or compositing point clouds from different sources into a shared scene, we can synthesize plausible redirected videos from the edited 4D geometry. This highlights the flexibility of explicit 4D representations over purely implicit approaches for scene manipulation.

%% file: sec/5_ablation.tex
\section{Ablation Study}
\label{sec:ablation}

\input{tab/ablation}

\begin{figure}[!t]
  \centering
  \includegraphics[width=\linewidth]{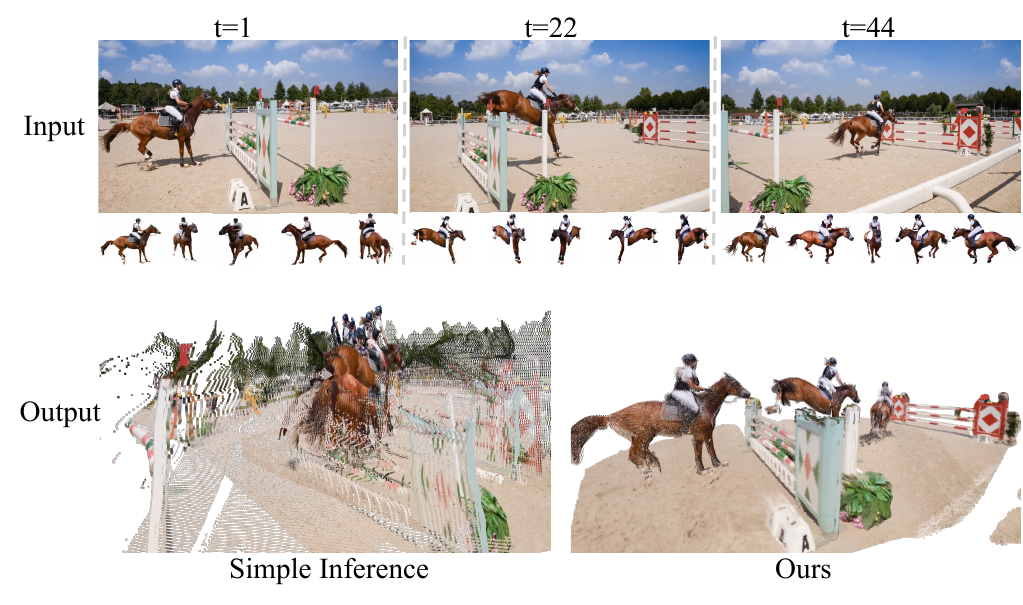}
  \caption{\textbf{Ablation: Simple Inference vs. Ours.} Directly combining multi-view images with source video leads to ghosting artifacts (left). Our decoupled strategy produces coherent 4D reconstruction (right).}
  \Description{Visual comparison showing ghosting artifacts from simple inference on the left versus clean geometry from our decoupled strategy on the right.}
  \label{fig:ablation}
\end{figure}

We conduct ablation studies on each key component under the same setting as our main experiments. Tab.~\ref{tab:ablation_results} and Fig.~\ref{fig:ablation} report the quantitative and qualitative results, respectively.

\paragraph{Simple Inference}
A straightforward baseline directly feeds the multi-view images together with the source video into a dynamic-aware feed-forward reconstruction network, bypassing our decoupled strategy. However, as shown in Fig.~\ref{fig:ablation}, this naive combination fails to produce a coherent 4D reconstruction: since the same object across frames exhibits similar appearance, the network collapses correspondences across time, producing misalignment and ghosting.

\paragraph{Without Multi-View Generation (MVG)}
We remove the multi-view generation step and rely solely on partial foreground point clouds for guidance. As shown in Tab.~\ref{tab:ablation_results}, performance degrades across all metrics, indicating that multi-view generation provides complete foreground geometry that is critical for stable synthesis under large viewpoint changes.

\paragraph{Without Kalman Filter (KF)}
We disable temporal smoothing. As discussed in Sec.~\ref{subsec:alignment}, monocular lifting produces frame-to-frame depth inconsistency that propagates into the aligned foreground trajectory. The performance drop upon removing the Kalman filter confirms its necessity for coherence.

%% file: tab/ablation.tex
\begin{table}[t]
    \centering
    \caption{\textbf{Ablation study.} We progressively add multi-view generation (MVG) and Kalman filter smoothing (KF) to evaluate each component's contribution. \textbf{Bold}: best.}
    \Description{An ablation table reporting DINO-SIM, CLIP-SIM, FID-V, and FVD-V for three configurations: a Baseline, Baseline plus multi-view generation (+MVG), and the full model that additionally adds Kalman filter smoothing (+KF). Adding each component progressively improves all four metrics, with the full model achieving the best score on every metric.}
    \small
    \setlength\tabcolsep{4pt}
    \resizebox{\columnwidth}{!}{%
    \begin{tabular}{l | c c c c}
        \toprule
        Method & DINO-SIM ($\uparrow$) & CLIP-SIM ($\uparrow$) & FID-V ($\downarrow \times 10^2$) & FVD-V ($\downarrow \times 10^3$) \\
        \midrule
        Baseline        & 0.58 & 0.81 & 1.9 & 4.1 \\
        + MVG           & 0.60 & 0.82 & 1.9 & 4.1 \\
        + KF (Full)     & \textbf{0.65} & \textbf{0.84} & \textbf{1.7} & \textbf{3.6} \\
        \bottomrule
    \end{tabular}}
    \label{tab:ablation_results}
\end{table}

%% file: sec/6_conclusion.tex
\section{Limitations}
\label{sec:limitations}
Our method has three main limitations.
First, the pipeline assumes a single dominant foreground object and a mostly static background. It can handle multiple objects when each is sufficiently observed over time, but complex interactions with heavy mutual occlusions or dynamic backgrounds remain challenging for current object-centric models.
Second, as a modular system built on off-the-shelf components, upstream errors in segmentation or multi-view synthesis can propagate to downstream stages. The modular design, however, enables independent component upgrades as stronger models emerge.
Third, the multi-stage pipeline prioritizes redirection quality over runtime efficiency, incurring substantial computational cost.

\FloatBarrier
\section{Conclusion}
\label{sec:conclusion}

We present \textbf{FreeOrbit4D}, a training-free framework for camera redirection from a single monocular video. By reconstructing a foreground-complete 4D proxy as geometric scaffolds for video generation, our method achieves SOTA fidelity, temporal coherence, and camera control under large viewpoint changes, while enabling applications such as appearance propagation and scene manipulation. Future work may explore integrating learned priors for improved robustness and leveraging the recovered 4D proxy for downstream tasks such as 4D asset creation and synthetic data generation.